\ifdefined\pdfoutput
\pdfoutput=1
\fi

\documentclass[11pt]{scrartcl}
\usepackage{array}
\usepackage{graphicx,multirow,listings,float}
\usepackage[labelformat=simple]{subcaption}

\usepackage[export]{adjustbox}
\usepackage{amsmath,amsthm,mathrsfs,amssymb}
\usepackage{bbm,bm,amsfonts}
\usepackage[mathscr]{eucal}
\usepackage{diagbox}
\usepackage{xcolor}
\usepackage{manyfoot,booktabs,textcomp}
\usepackage{algorithm,algorithmicx,algpseudocode}
\usepackage{authblk}
\usepackage{todonotes}
\usepackage[T1]{fontenc}
\usepackage[utf8]{inputenc}
\usepackage[a4paper,left=2cm,right=2cm]{geometry}
\usepackage[numbers]{natbib}
\usepackage{hyperref}
\usepackage[capitalize]{cleveref}
\DeclareOldFontCommand{\rm}{\normalfont\rmfamily}{\mathrm}
\ifdefined\DeclareUnicodeCharacter
\DeclareUnicodeCharacter{2011}{-}
\DeclareUnicodeCharacter{2013}{--}
\fi
\hypersetup{
  colorlinks=true,
  linkcolor=blue,
  urlcolor=blue,
  citecolor=blue
}

\newcommand{\bE}{\mathbb E}
\newcommand{\bS}{\mathbb S}
\newcommand{\bR}{\mathbb R}
\newcommand{\bZ}{\mathbb Z}
\newcommand{\bM}{\mathbb M}

\newcommand{\kq}{\mathfrak q}

\newcommand{\kC}{\mathfrak C}
\newcommand{\kF}{\mathfrak F}
\newcommand{\kH}{\mathfrak H}

\newcommand{\kM}{\mathfrak M}
\newcommand{\kU}{\mathfrak U}

\newcommand{\cR}{\mathcal R}
\newcommand{\cS}{\mathcal S}
\newcommand{\cU}{\mathcal U}

\newcommand{\cG}{\mathcal G}
\newcommand{\cH}{\mathcal H}
\newcommand{\cL}{\mathcal L}
\newcommand{\cC}{\mathcal C}
\newcommand{\cF}{\mathcal F}

\newcommand{\cD}{\mathcal D}

\newcommand{\cJ}{\mathcal J}
\newcommand{\cW}{\mathcal W}

\def\<{\langle} 
\def\>{\rangle}

\newcommand{\fM}{\mathbf M}

\newcommand{\fq}{\mathbf q}
\newcommand{\fp}{\mathbf p}
\newcommand{\fx}{\mathbf x}
\newcommand{\fV}{\mathbf V}

\newcommand{\dx}{\dot{x}}

\newcommand{\dfx}{{\dot{\mathbf{x}}}}
\newcommand{\dtheta}{{\dot{\theta}}}

\newcommand{\hfx}{{\hat{\mathbf{x}}}}
\newcommand{\hx}{{\hat{x}}}
\newcommand{\htheta}{{\hat{\theta}}}

\DeclareMathOperator\curl{curl}

\DeclareMathOperator\Id{Id}

\DeclareMathOperator\rn{\mathbf{n}}

\theoremstyle{plain}

\theoremstyle{definition}

\theoremstyle{remark}
\newtheorem{remark}{Remark}

\newcommand{\keywords}[1]{\par\noindent\textbf{Keywords:} #1}
\newenvironment{highlights}{\section*{Highlights}\begin{itemize}}{\end{itemize}}

\begin{document}

\title{Geodesics with Unified Tangent-constrained Priors and Curvature Regularization}

\author[1]{Chong Di\thanks{cdi@qlu.edu.cn}}
\author[2]{Li Liu\thanks{liuli184791@sjtu.edu.cn}}
\author[3]{Jinglin Zhang\thanks{jinglin.zhang@sdu.edu.cn}}
\author[4]{Zhenjiang Li\thanks{zhenjiangli1987@163.com}}
\author[5]{Da Chen\thanks{chenda@ceremade.dauphine.fr}}
\author[5]{Laurent D. Cohen\thanks{cohen@ceremade.dauphine.fr}}
\affil[1]{Shandong Artificial Intelligence Institute, Qilu University of Technology (Shandong Academy of Sciences), Jinan, Shandong, China}
\affil[2]{Yuanshen Rehabilitation Institute, Shanghai Jiao Tong University School of Medicine, Shanghai 200025, China}
\affil[3]{School of Control Science and Engineering, Shandong University, Jinan, Shandong, China}
\affil[4]{Department of Radiation Oncology, Shandong Cancer Hospital and Institute, Shandong First Medical University, Shandong Academy of Medical Sciences, Jinan, Shandong, China}
\affil[5]{CEREMADE, Universit\'e Paris Dauphine, Universit\'e-PSL, CNRS, UMR 7534, Paris 75775, France}
\renewcommand\Authfont{\normalsize}
\renewcommand\Affilfont{\small}
\setlength{\affilsep}{0.35em}
\date{}
\maketitle

\begin{abstract}
Curvature-penalized geodesic models have proven their effectiveness in image segmentation by computing globally optimal curves. Unfortunately, these models remain susceptible to shortcuts when delineating objects with complex shapes and image intensity distributions, as they lack mechanisms to enforce shape-aware tangent constraints. To address this limitation, we propose a unified geodesic framework that integrates tangent-constrained priors with curvature penalization. The key idea is to formulate tangent admissibility directly within the orientation-lifted space, where path tangents are restricted to spatially varying angular sectors derived from intrinsic shape representatives (ISR) such as skeletons or interior landmarks. This formulation gives rise to a family of tangent-constrained Finslerian metrics, extending the classical curvature-penalized geodesic models while enforcing mandatory tangent constraints. The resulting Hamilton-Jacobi-Bellman (HJB) partial differential equations (PDEs) admit efficient numerical solutions via variants of the fast marching method, preserving the single-pass computational complexity. Experiments on synthetic, natural, and medical images demonstrate that the proposed geodesic framework indeed improves robustness against weak boundaries and topological shortcuts, yielding segmentation results with enhanced shape fidelity compared to existing geodesic models.
\end{abstract}

\keywords{Circular geodesic, fast marching method, image segmentation, tangent-constrained prior, curvature penalization}

\begin{highlights}
    \item \textbf{Unified tangent-constrained curvature-penalized geodesic models:} We introduce a family of tangent-constrained Finslerian metrics that extend classical curvature-penalized geodesic models (including Reeds--Shepp, Dubins, Euler--Mumford elastica and curvature prior elastica models). The proposed formulation enforces hard tangent constraints while remaining fully compatible with curvature regularization.
    
    \item \textbf{HJB PDE formulation with efficient numerical solvers:} The resulting Hamilton--Jacobi--Bellman equations can be efficiently solved using variants of Hamiltonian fast marching methods, preserving single-pass computational complexity and enabling practical application to large-scale image segmentation tasks.
    
    \item \textbf{Practical segmentation pipeline with deep-learning-based initialization:} We present an automatic segmentation framework that leverages neural network pre-segmentation to extract intrinsic shape representatives, which are then used to construct tangent priors and initialize circular geodesic models. Extensive experiments on synthetic, natural, and medical images demonstrate that the proposed approach significantly improves robustness to weak boundaries and complex object geometries, outperforming classical geodesic models as well as representative graph-based and learning-based baselines.
\end{highlights}

\section{Introduction}
\label{sec_introduction}

\begin{figure}[t]
\centering{\includegraphics[width=0.7\linewidth]{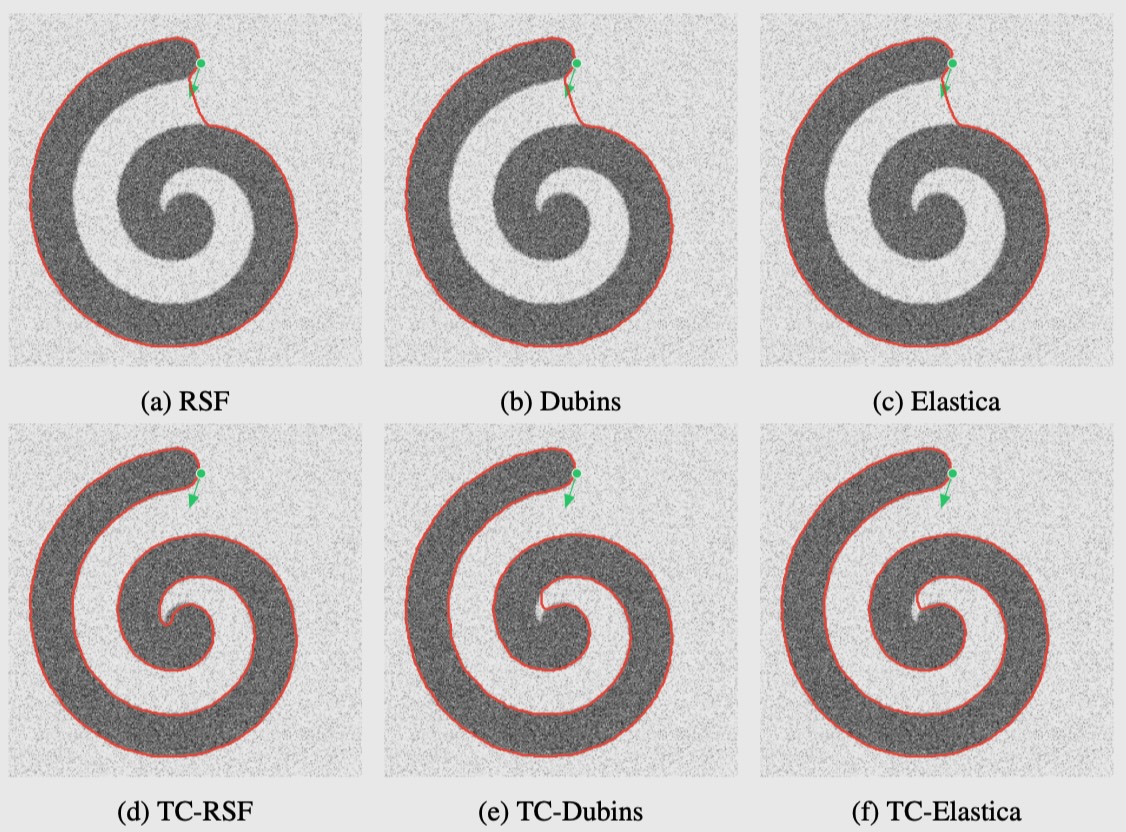}}
\caption{Comparison of segmentation results on a spiral shape between the classical curvature-penalized geodesic models and the proposed tangent-constrained models. (a) to (c) Segmentation results generated using the RSF, Dubins, and EM elastica models, respectively. (d) to (f) Segmentation results generated by the TC-RSF, TC-Dubins, and TC-Elastica models, respectively.}
  \label{fig_introduction}
\end{figure}

Image segmentation remains a fundamental challenge in computer vision and medical imaging~\citep{yang2023active,minaee2021image,azad2024medical,yuan2025saliency,huang2025rethinking,zheng2025BGPSeg}. Despite significant advances, existing segmentation approaches are still struggling with weak boundaries, noise, complex object geometries and image intensity distributions. Both global and local shape constraints have proven essential for achieving suitable segmentations across various scenarios~\citep{cremers2007review}. In practice, shape constraints serve as effective geometry priors in deep learning-based models, energy minimization-based geometric models and hybrid frameworks combining both paradigms.

Energy minimization is a flexible framework for addressing image segmentation, capable of encoding various types of image features, geometric regularization, and shape priors into a unified objective functional. Existing energy minimization models can be broadly divided into two categories: graph-based models (e.g.,~\cite{boykov2006grapy,krahenbuhl2011efficient,zheng2015conditional,veksler2024sparse,wu2022balanced}) and variational models~(e.g.,~\cite{mumford1989optimal,kass1988snakes,li2010distance,cohen1993finite}). Graph-based segmentation models cast the problem as a discrete optimization problem, which can be solved efficiently using algorithms such as graph cuts, random walks or power watershed~\citep{grady2006random,boykov2004experimental,couprie2010power}. However, they typically approximate geometric quantities using finite-order clique potentials defined over a discrete lattice, often leading to metrication artifacts and limiting the incorporation of high-order geometric priors~\citep{el2016contrast,ulen2015torsion,schoenemann2012linear}. In contrast, 
variational models address segmentation within a PDE framework, providing a continuous, geometry-preserving representation that naturally accommodates curvature, other high-order geometric regularization terms and shape priors, without being restricted to discrete structures. The proposed segmentation method falls into the latter category, leveraging the PDE-based formulation to integrate curvature penalization and local shape constraints in a principled manner.

The minimal path model~\citep{cohen1997global} was first introduced to obtain the global minimization of a path energy associated with a suitable isotropic Riemannian metric, establishing the connection between the PDE and the differential geometry. It was originally investigated to track image edges and tubular structures. These applications have inspired a series of works to address the limitations of this original model, such as the integration of anisotropic and/or asymmetric metrics~\citep{chen2019minimal,chen2021generalized,chen2021geodesic,liao2022progressive,liao2018progressive,benmansour2011tubular}. Among those geodesic approaches, a significant research line is the curvature-penalized models, which incorporate second-order curvature terms for regularization, and have demonstrated promising results in several practical applications~\citep{duits2018optimal,mirebeau2018fast,van2024geodesic,chen2023computing,chen2017global}. These models must address a fundamental issue of incorporating a second-order curvature regularization within an inherently first-order HJB PDE framework. The orientation-lifting strategy provides an elegant resolution by introducing an auxiliary angular dimension, thereby expressing path curvature as a ratio of first-order derivatives. Existing second-order geodesic models either employ curvature as a regularization term or leverage it for tracking circular minimal paths with a convexity shape prior. However, they cannot impose mandatory directional constraints on the path tangents to accommodate more flexible, spatially varying shape constraints. 
Fig.~\ref{fig_introduction} illustrates a fundamental limitation inherent to classical curvature-penalized models when applied to complicated segmentation scenarios. We consider examples of segmenting a spiral using three representative models: the Reeds-Shepp forward (RSF) model~\citep{duits2018optimal}, the Dubins model~\citep{mirebeau2018fast} and the Euler-Mumford elastica model~\citep{chen2017global}. 
As shown in Figs.~\ref{fig_intro-rsf} to~\ref{fig_intro-em}, all three models produce geodesics that take undesirable shortcuts. This shortcutting behavior arises because these models lack any mechanism to enforce directional consistency with the underlying shape geometry. In contrast, Figs.~\ref{fig_intro-tc-rsf} to~\ref{fig_intro-tc-em} present the segmentation results obtained by integrating tangent-constrained priors into the same curvature-penalized geodesic frameworks. The resulting models\footnote{Corresponding to their classical counterparts, the proposed models are referred to as the TC-RSF, TC-Dubins and TC-Elastica, respectively.} successfully delineate the complete spiral boundary without shortcuts, demonstrating that the proposed tangent constraints effectively resolve this limitation while preserving the benefits of curvature regularization. 

In this work, we propose a unified geodesic framework that integrates tangent-constrained priors and curvature penalization within a single PDE formulation. The tangent constraints are derived from the ISR and formulated as orientation-dependent costs in a high-dimensional space, enabling the use of efficient single-pass Hamiltonian Fast-Marching methods~\citep{mirebeau2018fast,mirebeau2019hamiltonian} or the GPU-accelerated solvers for the HJB PDE~\citep{mirebeau2023massively}. Compared to the convexity shape prior, the proposed tangent-constrained priors are less restrictive and thus applicable to a broader range of segmentation scenarios. To the best of our knowledge, the most relevant works to our model are the star convexity-embedded approaches~\citep{gulshan2010geodesic,veksler2008star,isack2016hedgehog,liu2022deep,zhao2025convex}. Among these, the star shape constraint is either incorporated into graph-based models, or integrated with neural networks to guide the segmentation output, both of which have achieved notable success. However, the graph-based models are susceptible to metrication errors and anisotropic bias, while neural network-based approaches rely on soft relaxations of geometric constraints, offering no theoretical guarantee that the output satisfies the star convexity property. Moreover, neither class of models incorporates curvature penalization, which plays a crucial role in the presence of strong noise, weak edges and complex image content. This provides the motivation for this work, where the main contributions are two-fold:
\begin{itemize}
	\item We propose a unified HJB PDE framework for computing tangent-constrained and curvature-penalized circular optimal paths of minimum arrival time, where the tangent priors are derived from the ISR that serves as local shape constraints. We present an efficient method that transforms the vector-valued tangent priors as a scalar-valued field over the orientation-lifted space, allowing to invoke the well-established numerical solvers (e.g. Hamiltonian fast marching method~\cite{mirebeau2018fast} or fast iterative method~\cite{mirebeau2023massively}) for computing satisfactory minimal paths.
	
	\item A practical pipeline that leverages deep learning-based pre-segmentation to extract the ISR (e.g., skeletons or interior landmarks) and to provide automatic initialization for image segmentation based on the proposed geodesic models. This enables automatic segmentation while ensuring that the resulting circular geodesics encode both image features and tangent constraints, thereby effectively preventing unexpected shortcuts in challenging scenarios.
\end{itemize}

The manuscript is organized as follows. Section~\ref{sec_Background} reviews the curvature-penalized minimal path models based on the static HJB PDE framework. Section~\ref{sec_MainConstribution} presents the core of the proposed model: the unified tangent-constrained and curvature-penalized geodesic framework. Section~\ref{sec_AppSeg} describes the applications of the introduced geodesic models to image segmentation. Section~\ref{sec_Exp} reports experimental results and Section~\ref{sec_Conclusion} concludes the paper.

\section{Curvature-penalized Minimal Paths}
\label{sec_Background}
In this section, we revisit the curvature-penalized minimal path models upon the HJB PDE framework, which are posed in an orientation-lifted space~\citep{chen2017global,mirebeau2018fast,duits2018optimal,chen2023computing}.

\subsection{Orientation Lifting for Path Tangents and Curvature}
To represent curvature within the energy minimization framework, the open bounded and connected domain $\Omega \subset \bR^{2}$ is lifted to an orientation-lifted space $\bM := \Omega \times \bS^{1}$, where $\bS^1 := \bR/2\pi\bZ$ is an angle space with period boundary condition. Any point $\fx \in \bM$ is a pair $(x, \theta)$ consisting of a spatial coordinate $x \in \Omega$ and an angular coordinate $\theta \in \bS^{1}$. The tangent space at any point $\fx$ is given by $\bE := \bR^{2} \times \bR$, with tangent vectors denoted by $\dfx = (\dx, \dtheta)$.

A smooth planar curve $\gamma:[0,1] \to \Omega$ with a non-vanishing velocity (i.e., $\|\gamma^\prime(\cdot)\|\neq 0$) can be lifted to a curve $\Gamma:[0,1] \to \bM$, defined as $\Gamma = (\gamma, \eta)$. The angular component $\eta(t) \in \bS^{1}$ represents the orientation of the curve $\gamma$ for any $t\in[0,1]$. Consequently, the tangent vector of the planar curve is constrained to be aligned with its orientation:
\begin{equation}
\label{eq:first_order_gamma}
\gamma^\prime(t) = \|\gamma^\prime(t)\| \rn(\eta(t)),
\end{equation}
where $\rn(\theta) = (\cos\theta, \sin\theta)$ denotes the unit vector associated with the angle $\theta$. The first-order derivative of the orientation-lifted curve is then given by $\Gamma^\prime(t)= \big(\gamma^\prime(t), \eta^\prime(t)\big),\,\forall t\in[0,1]$.

This orientation-lifted representation allows for a direct formulation of the curvature $\kappa(t)$ of $\gamma$. It is defined as the ratio of two first-order terms
\begin{equation}
  \kappa(t) := \frac{\eta^\prime(t)}{\|\gamma^\prime(t)\|},
\end{equation}
 allowing to incorporate the curvature terms into the first-order static HJB PDE framework.

\subsection{Curvature-penalized Geodesic Models}
The bending energy of a regular curve $\gamma : [0,1] \to \Omega$ is formulated to include a penalty on its curvature, which generalizes the Euclidean curve length as follows
\begin{equation}
\label{eq:energy}
\int_{0}^{1} 
\psi(\gamma(t), \eta(t))\;
\cC(\xi\kappa(t))
\|\gamma^\prime(t)\|\, dt,
\end{equation}
where $\psi:\bM \to \bR^{+}$ is an image data-driven cost function, $\cC:\bR \to [0,\infty]$ is a cost function of curvature dependent on the specific geodesic models\footnote{In Appendix~\ref{appendix_CurvatureModels}, we show different types of  curvature-penalized geodesic models and their corresponding curvature cost functions $\cC$.}, and $\xi\in\bR^+$ is a constant modulating the strength of curvature penalization. 
The term $\eta(t)$ denotes the turning angle of $\gamma(t)$ satisfying the assumption~\eqref{eq:first_order_gamma}. 

In the orientation-lifted space, this energy corresponds to the length of the lifted path $\Gamma(t) = (\gamma(t), \eta(t))$:
\begin{equation}
  \cL(\Gamma) := \int_{0}^{1} 
  \psi(\Gamma(t))\cF(\Gamma(t),\Gamma^\prime(t)) \, dt,
\end{equation}
 where $\cF:\bM \times \bE \to [0,\infty)$ is an orientation-lifted Finsler metric. For a point $\mathbf{x}=(x,\theta) \in \bM$ and a tangent vector $\dot{\mathbf{x}}=(\dot{x},\dot{\theta}) \in \bE$, the metric is defined as:
 \begin{equation}
 \label{eq:metric}
 \cF(\mathbf{x}, \dot{\mathbf{x}}) =
\begin{cases}
\cC\bigl(\xi \dtheta/\|\dx\|\bigr)\|\dx\|, & \text{if~} \dx= \rn(\theta)\|\dx\|, \\
\infty, & \text{otherwise}.
\end{cases}
\end{equation}
The infinite penalty enforces the constraint that the spatial velocity $\dx$ must be aligned with the orientation vector $\rn(\theta)$.

The geodesic path between a source point $\fp\in \bM$ and a target point $\fx\in\bM$ is found by computing the minimal action, also known as the geodesic distance map $\cU_{\fp}$:
\begin{equation}
\label{eq:distance_map}
 \cU_{\mathbf{p}}(\mathbf{x}) =
\inf_{\Gamma \in \mathrm{Lip}([0,1], \bM)}
\left\{
\cL(\Gamma) \,;\,
\Gamma(0) = \mathbf{p},\ \Gamma(1) = \mathbf{x}
\right\}.
\end{equation}
This distance map $\cU_{\mathbf{p}}$ is the unique viscosity solution of a static Hamilton-Jacobi equation. The equation is defined in terms of the Hamiltonian $\cH$ associated with the metric $\cF$:
\begin{equation}
\label{eq_HJBPDE_Curvature}
 \cH(\fx,d\cU_{\fp}(\fx))= \frac{1}{2}\,\psi(\fx)^{2},\quad\forall \fx \in \bM \backslash \{\fp\},
\end{equation}
subject to $\cU_{\mathbf{p}}(\mathbf{p})=0$ and an outflow boundary condition on $\partial \bM$. Here, $d\cU_{\mathbf{p}}$ represents the differential of the geodesic distance map $\cU_\fp$. The Hamiltonian $\cH$ is obtained through the Legendre-Fenchel duality of the metric $\cF$:
\begin{equation}
\label{eq:hamiltonian}
 \cH(\fx,\hfx):= \sup_{\dfx \in \bE}
\left\{
\left\langle \hat{\mathbf{x}}, \dot{\mathbf{x}} \right\rangle
- \frac{1}{2}\,\cF(\mathbf{x}, \dot{\mathbf{x}})^{2}
\right\},
\end{equation}
for any $\fx=(x,\theta) \in \bM$ and any $\hfx=(\hx,\htheta) \in \bR^{2} \times \bR$.

Once the geodesic distance map $\cU_{\fp}$ is computed, the geodesic path $\cG_{\fp,\fx}$ from the source $\fp$ to any target point $\fx \in \bM$ can be recovered by a backtracking scheme. This is implemented by solving a gradient descent ordinary differential equation (ODE) backward in time from the target, reading for any $t \in (0, T]$
\begin{equation}
  \cG'(t) = \mathbf{V}\left(\cG(t)\right),\quad \mathbf{V}(\fx)= \partial_2\cH(\fx,d\cU_{\fp}(\fx))
\end{equation}
till the source point $\fp$ is reached, i.e. $\cG(T)=\fp$ is achieved, where $T:=\cU_{\mathbf{p}}(\mathbf{x})$ is the arrival time and where $\fV$ is the geodesic flow. Finally, the target geodesic path $\cG_{\fp,\fx}$ can be obtained by reparameterizing $\cG$ such that $\cG_{\fp,\fx}(0)=\fp$ and $\cG_{\fp,\fx}(1)=\fx$. Note that $\partial_2\cH(\fx,\hfx):=\partial\cH(\fx,\hfx)/\partial\hfx$ denotes the partial differential of the Hamiltonian $\cH$ with respect to its second argument.

\section{A HJB PDE Framework for Computing Curvature-penalized Geodesics with Tangent-Constrained Priors}
\label{sec_MainConstribution}
In this section we present a unified minimal path framework that incorporates both path tangent constraints and curvature penalization. We begin by introducing the formulation of tangent constraints interpreted in the orientation-lifted space $\bM$, followed by the development of three distinct geodesic models that integrate these constraints with curvature penalization. 

\subsection{Constructing the Tangent-Constrained Priors}
In image segmentation, a crucial challenge lies at the efficient representation of object shapes. Among various representations, the shape skeleton or interior landmark points are known to offer compact yet expressive descriptions. Hereafter, the skeleton or the interior landmark points are referred to as ISR. It encodes the internal geometric organization of a region, thus can be taken as an important cue to derive shape appearance features. In order to extend the effective scope to the whole domain $\Omega$, a common way is to construct a distance map emanating from the skeleton structure~\citep{veksler2008star,gulshan2010geodesic,liu2022deep,isack2016hedgehog}. In the remainder of this subsection, we discuss the computation for the geometric constraints using ISR. 

\subsubsection{Distances from the ISR}
Let $\Im\subset \Omega$ denote the ISR, from which one can compute a weighted distance map $\cD_{\Im}$ using the Eikonal PDE~\citep{cohen1997global,mirebeau2019riemannian}
\begin{equation}
\label{eq_PriorHJB}
\begin{cases}
\tilde\cH(\fx, d\cD_{\Im}(\fx))=\frac{1}{2},&\forall \fx\in\Omega\backslash\Im, \\
\cD_{\Im}(\fx)=0,&\forall\fx\in\Im	
\end{cases}
\end{equation}
where $\tilde\cH$ is a Hamiltonian. 
In particular for the Riemannian case, the Hamiltonian $\tilde\cH$ can be expressed as 
\begin{equation}
\tilde\cH(\fx,\hfx)=\left\langle\hfx ,\kM(\fx)\hfx\right\rangle,
\end{equation}
where $\kM$ is a tensor field such that $\kM(\fx)$ is a symmetric definite positive matrix of size $2\times2$, and where $\langle\cdot,\cdot\rangle$ is the standard Euclidean scalar product in $\bR^2$. The tensor field $\kM$ can be derived from the image data, for instance using image gradients~\citep{gulshan2010geodesic}, or from the shape of prescribed segmentation mask. In particular, the geodesic distance map $\cD_\Im$ can be reduced to a Euclidean distance map by setting $\kM\equiv\Id$, where $\Id$ is the identity of size $2\times2$.
The Eikonal PDE~\eqref{eq_PriorHJB} can be efficiently solved using the fast marching method~\citep{sethian1996fast,mirebeau2014anisotropic,mirebeau2019riemannian}.

 Let $\tilde\kq:\Omega\to\bR^2$ be a \emph{unit} vector field subject to $\|\tilde\kq\|\equiv1$, which can be naturally defined using the gradient of the distance map $\cD_{\Im}$
 \begin{equation}
 \tilde\kq(\fx):=
 \begin{cases}
\displaystyle \frac{\partial_2\tilde\cH(\fx,d\cD_{\Im}(\fx))}{\|\partial_2\tilde\cH(\fx,d\cD_{\Im}(\fx))\|},&\forall\fx\in\Omega\backslash\Im,\\
 \mathbf{0},&\text{otherwise}	
 \end{cases}	
 \end{equation}
which leads to a rotated unit vector field $\kq$ such that
\begin{equation}
\label{eq_tangentPriors}
 \kq(\fx)=\fM(\pi/2)\;\tilde\kq(\fx),	
 \end{equation}
where $\fM(\pi/2)$ is a counter-clockwise rotation matrix with respect to an angle $\pi/2$. Hereafter, we refer to the vector field $\kq$ as the tangent priors. 

\subsubsection{Tangent-constrained priors in the orientation-lifted space $\bM$}
\label{sec_tangent_constraint}
We introduce a first-order constraint to limit the path tangents. For a regular curve $\gamma:[0,1]\to\Omega$, we assume that its tangent $\gamma^\prime(t)$ satisfies the following acute alignment inequality 
\begin{equation}
\label{eq_TangentConstraint}
\bigl\langle\gamma^\prime(t),\kq(\gamma(t)) \bigr\rangle \geq \rho(\gamma(t))\|\gamma^\prime(t)\|,\quad \forall t\in[0,1],
\end{equation}
 where $\rho:\Omega\to[0,1]$ is a scalar-valued function. The inequality in~\cref{eq_TangentConstraint} imposes a mandatory constraint to the path tangent, i.e., the angle between the vectors $\gamma^\prime(t)$ and $\kq(\gamma(t))$ for any $t\in[0,1]$ is required to be \emph{acute}, since the values of $\rho$ are limited to the range $[0,1]$. 
 
The orientation lifting of a regular curve $\gamma$ states that the tangent $\gamma^\prime(t)$ can be characterized by its turning angles $\eta$ via ~\cref{eq:first_order_gamma}. In this case, the inequality~\eqref{eq_TangentConstraint} can be rewritten as 
\begin{equation}
\label{eq_OLAcuteConstraint}
\langle\rn(\eta(t)),\kq(\gamma(t)) \rangle\geq \rho(\gamma(t)).
\end{equation}
It allows to define a binary-valued function $\kC:\bM\to\{0,\infty\}$ encoding the inequality~\eqref{eq_OLAcuteConstraint}
\begin{equation}
\label{eq_AcuteConstraint}
\kC_{\rho,\kq}(\fx)=
\begin{cases}
	0,&\langle\rn(\theta),\kq(x) \rangle\geq \rho(x)\\
	\infty,&\text{otherwise}
\end{cases}
\end{equation}
for any point $\fx=(x,\theta)\in\bM$, which can be used to build a HJB PDE framework for computing minimal paths with both curvature penalization and tangent constraint.

\subsection{Tangent-constrained Curvature-penalized Geodesic Models}
\label{sec:tangent_constrained_models}
In this section, we introduce the core of this work: a HJB PDE framework for computing minimal paths with both curvature penalization and tangent constraint, based on the acute alignment constraint defined in~\cref{eq_TangentConstraint,eq_OLAcuteConstraint}. 
In conjunction with $\kC_{\rho,\kq}$ formulated in~\cref{eq_AcuteConstraint}, we introduce a new type of geodesic metric $\kF_{\rho,\kq}:\bM\times\bR^3\to[0,\infty]$, encoding the tangent constraint and reading for any point $\fx=(x,\theta)$ and for any vector $\dfx=(\dx,\dtheta)$ as
\begin{align}
\label{eq_ConstrainedMetric1}
\kF_{\rho,\kq}(\fx,\dfx)&=\begin{cases}
\left(\cC\bigl(\xi \dtheta/\|\dx\|\bigr)+\kC_{\rho,\kq}(\fx)\right)\|\dx\|, & \text{if~} \dx= \rn(\theta)\|\dx\|,\\
\infty,& \text{otherwise}
\end{cases}\\
\label{eq_ConstrainedMetric2}
&=\begin{cases}
\cF(\fx,\dfx),&\text{if~}\langle\rn(\theta),\kq(x) \rangle\geq \rho(x)\\
\infty,&\text{otherwise},
\end{cases}
\end{align}
where $\cF$ is a curvature-penalized geodesic metric dependent on the models, as described in Appendix~\ref{appendix_CurvatureModels}.
For any point $\fx=(x,\theta)$, the geodesic metric $\kF_{\rho,\kq}(\fx,\dfx)$ is assigned the value $\infty$ if the prescribed acute alignment constraint is not satisfied.

The computation for minimal paths under the introduced metric $\kF_{\rho,\kq}$ can be performed within the HJB PDE framework. The corresponding Hamiltonian of the metric $\kH_{\rho,\kq}:\bM\times\bR^3\to[0,\infty]$ can be constructed using $\cH$ of $\cF$, see~\cref{eq:hamiltonian}. This is to say
\begin{equation}
\kH_{\rho,\kq}(\fx,\hfx):=\cH(\fx,\hfx),\quad \text{if~}\langle\rn(\theta),\kq(x) \rangle\geq \rho(x),
\end{equation}
for any point $\fx=(x,\theta)\in\bM$ and any co-vector $\hfx=(\hx,\htheta)$. 

Given a source point $\fp$ and a target point $\fq$ located in $\bM$, the minimal path $\cG_{\fp,\fq}$ between them can be generated using a geodesic distance map $\kU_{\fp}$ by solving a HJB PDE associated with the Hamiltonian $\kH_{\rho,\kq}$, i.e. the distance map $\kU_{\fp}$ satisfies that $\kU_\fp(\fp)=0$ and otherwise
\begin{equation*}
\kH_{\rho,\kq}(\fx,d\kU_\fp(\fx))=\frac{1}{2}\psi(\fx)^2,\quad \forall\fx\in\bM\backslash\{\fp\},
\end{equation*}
whose solution can be obtained by the Hamiltonian fast marching method~\citep{mirebeau2018fast}.

Alternatively, we introduce here a practical way for computing the geodesic distance map $\kU_\fp$ by reconstructing its supporting domain using the inequality~\eqref{eq_OLAcuteConstraint}:
\begin{equation}
\bM_{\kq,\rho}:=\bigl\{\fx=(x,\theta)\in\bM~|~\langle\rn(\theta),\kq(x) \rangle\geq \rho(x)\bigr\}.	
\end{equation}
Then we estimate the viscosity solution to the following HJB PDE using the domain $\bM_{\kq,\rho}$
\begin{equation}
\label{eq_HJB_ours}
\begin{cases}
\cH(\fx,d\kU_\fp(\fx))=\frac{1}{2}\psi(\fx)^2,&\quad \forall\fx\in\bM_{\kq,\rho}\backslash\{\fp\},\\
\kU_\fp(\fp)=0.&
\end{cases}
\end{equation}

In this work, we use the method introduced in~\citep{chen2023geodesic} for computing the data-driven cost $\psi$, which encodes both of the image gradients and the image region-based appearance features. 

\begin{remark}
It is worth noting that anisotropic geodesic metrics such as the asymmetric quadratic metric~\citep{chen2021generalized} and the anisotropic Riemannian metric~\citep{benmansour2011tubular,law2013gradient}) can also bias the path tangents toward a prescribed vector field. However, these metrics fundamentally differ from the proposed tangent constrained framework in two critical aspects: Firstly, anisotropic metrics only work with first-order geometry and thus cannot incorporate second-order curvature penalization within their formulation. Secondly, while anisotropic metrics impose soft directional preferences through weighted costs, they do not enforce hard constraints on path tangents. As a consequence, the resulting geodesics may still deviate from the admissible angular sectors, offering no theoretical guarantee that the tangent-constrained priors are strictly satisfied.
\end{remark}

\begin{figure*}[t]
\centering
 \includegraphics[width=0.8\linewidth]{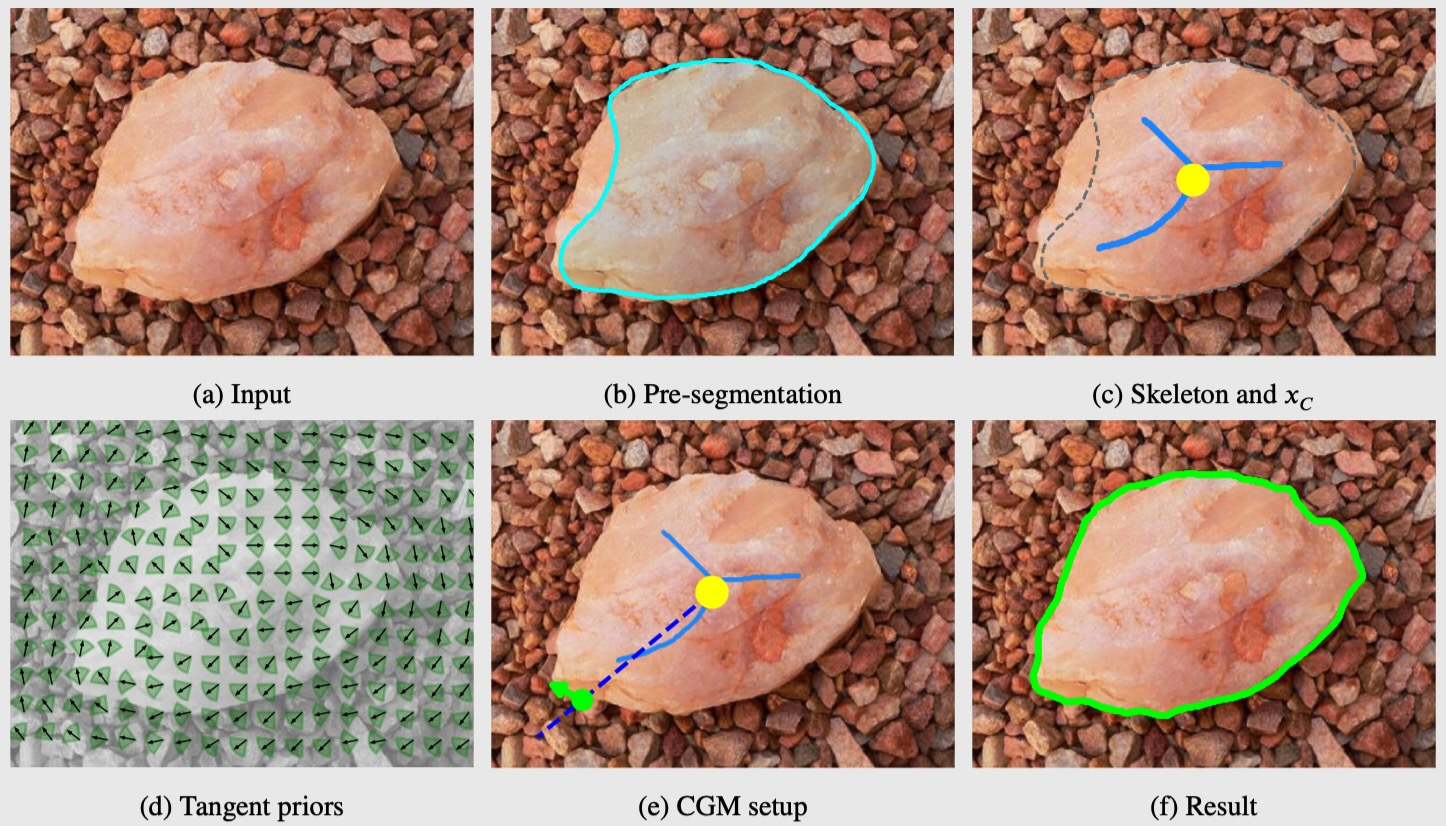}
\caption{Overview of the proposed tangent-constrained curvature-penalized segmentation pipeline. 
  (a)~Input image. 
  (b)~Pre-segmentation mask obtained by the DeepLabv3+ network. 
  (c)~Extraction of the ISR: the morphological skeleton~$\mathcal{S}$ (dodger blue) and the centroid~$x_C$ (yellow point). 
  (d)~Tangent priors~$\mathfrak{q}$ derived from the euclidean distance map associated to the region skeleton. 
  (e)~Initialization of the circular geodesic model: the seed point~$z$ (yellow dot), source point~$\mathbf{p}$ (green dot with arrow), and ray cut (blue dashed line). 
  (f)~Final segmentation result generated by the proposed TC-RSF model.}
  \label{fig:pipeline}
\end{figure*}

\section{Applications to Robust Image Segmentation}
\label{sec_AppSeg}
In this section, we present a practical method that applies the proposed geodesic framework with tangent-constrained priors and curvature penalization to image segmentation. While the framework in~\cref{sec_MainConstribution} supports interactive segmentation where users manually specify the ISR, we develop here a fully automatic approach by leveraging deep learning-based pre-segmentation to extract the ISR.

The proposed segmentation method mainly consists of three subsequent stages: (1) generating the ISR from neural network pre-segmentation (\cref{sec:pre-segmentation}), (2) constructing the image data-driven cost $\psi$ (\cref{subsec_DataCost}), and (3) computing the final segmentation contours via the adaptive circular geodesic models~\citep{appleton2005globally,liu2024grouping,chen2023geodesic} (\cref{subsec_Circular}). Fig.~\ref{fig:pipeline} illustrates the complete segmentation method.


\subsection{Generation of the ISR}
\label{sec:pre-segmentation}

In our work, the ISR are first constructed to initialize the tangent-constrained geodesic model. These representatives are produced through a pre-segmentation procedure using a pre-trained neural network. While various architectures such as PolarMask~\citep{xie2021polarmask++} and other prevalent segmentation models~\citep{minaee2021image} are suitable, we employ the DeepLabv3+~\citep{chen2018encoder} model. Its encoder-decoder architecture equipped with atrous spatial pyramid pooling effectively captures multi-scale context and preserves boundary details, thereby ensuring the generation of topologically consistent initial masks for the subsequent extraction of ISR.

From the probability map output by the network, we obtain a binary segmentation mask by thresholding. We then extract the largest connected component to represent the object of interest. The centroid of this region, denoted as $x_C$, is computed to serve as the reference center. Additionally, we extract the morphological skeleton $\cS$ of the binary mask, which provides a topological representation of the object's shape and is used to define the tangent constraint field.

\subsection{Data-driven Cost from Image Features}
\label{subsec_DataCost}
As discussed in~\cref{sec_MainConstribution}, the HJB PDE~\eqref{eq_HJB_ours} of the proposed model involves two components, the tangent-constrained priors and the image data-driven cost $\psi$. Following the literature~\citep{chen2023geodesic}, the data-driven cost $\psi$ in our work integrates both the image gradients-based and region-based appearance features. 
In particular, the region-based appearance features are dependent on the evolving shape~\citep{chen2023geodesic,chen2024region}. We denote by $U\subset\Omega$ a tubular neighbourhood that surrounds the boundary of the evolving shape.
For each point $\fx = (x, \theta) \in\bM$, the data-driven cost function, defined in an orientation-scores manner, is formulated as 
\begin{equation}
\psi(x,\theta)=
\begin{cases}
\exp\left(\alpha\,\tilde\psi(x,\theta)\right),&\forall x\in U\\
\infty,&\text{otherwise},	
\end{cases}	
\end{equation}
where $\alpha>0$ is a weighting parameter. The function $\tilde{\psi}$ combines both types of feature as follows
\begin{equation}
\label{eq_SubCosts}
\tilde\psi(x,\theta)=\psi_{\rm grad}(x,\theta)+\frac{\mu\langle\omega(x),\rn(\theta)\rangle}{\max_{y\in U}\|\langle\omega(y),\rn(\theta) \rangle\|},
\end{equation}
where $\psi_{\rm grad}$ is the orientation score from the image gradients and $\omega$ is a vector field of sufficiently small magnitude, respectively formulated in Appendices~\ref{appendix_OSGrad} and~\ref{appendix_RegionalScores}.
The first term in Eq.~\eqref{eq_SubCosts} captures the edge probability map through the structure tensor $\cW$, while the second term encodes regional preference via the vector field $\omega$, with $\mu>0$ balancing their relative contributions. Note that the vector field $\omega$ and the tubular neighbourhood will be updated according to the evolving shapes. 

%

\subsection{Final Segmentation with Circular Geodesic Model}
\label{subsec_Circular}
The circular geodesic model (CGM) is a practical method to extract closed planar geodesic curves within the image domain $\Omega$, which provides a feasible geodesic-based solution to the task of image segmentation.
In contrast to the original interactive context, where a point $\fp$ on the target boundary and its tangent $\theta_p$ are user-defined, our implementation automates this initialization using the pre-segmentation results.

The seed point $z$ required by the CGM is set to the centroid $x_C$ provided by the pre-segmentation stage. The source point $\fp$ on the target boundary is automatically selected as the point on the contour of the pre-segmentation mask that is farthest from the centroid $z$. This selection strategy typically places $\fp$ at an extremity of the object, facilitating the robust propagation of the geodesic front. The corresponding initial orientation $\theta_p$ is estimated from the tangent vector of the contour at $\fp$.

A ray line connecting the centroid $z$ and the source point $\fp$ is introduced as a cut to locally disconnect the domain, transforming the problem into finding a path from $\fp$ to itself in the cut domain. The skeleton $\mathcal{S}$ is also utilized as a set of internal obstacles to ensure the generated curve properly encloses the object's core structure. Minimal paths are then computed using the Hamiltonian fast marching method over the orientation-lifted domain.

%

\section{Experimental Results}
\label{sec_Exp}

In this section, we evaluate the performance of the proposed geodesic models that unify tangent-constrained priors with curvature penalization. We first conduct a study to validate the effectiveness of the tangent-constrained priors when integrated with different curvature penalization metrics, including the RSF, Dubins, and Elastica models. Next, we demonstrate how these priors improve segmentation accuracy when using neural network pre-segmentation results as initialization. Finally, we compare the geodesic models against two state-of-the-art methods that also benefit from pre-segmentation: a graph-based model with convexity constraints (Graph-Convexity) and the Segment Anything Model (SAM).

\subsection{Study on Tangent-Constrained Priors}

\begin{figure}[t]
\centering
\includegraphics[width=0.6\linewidth]{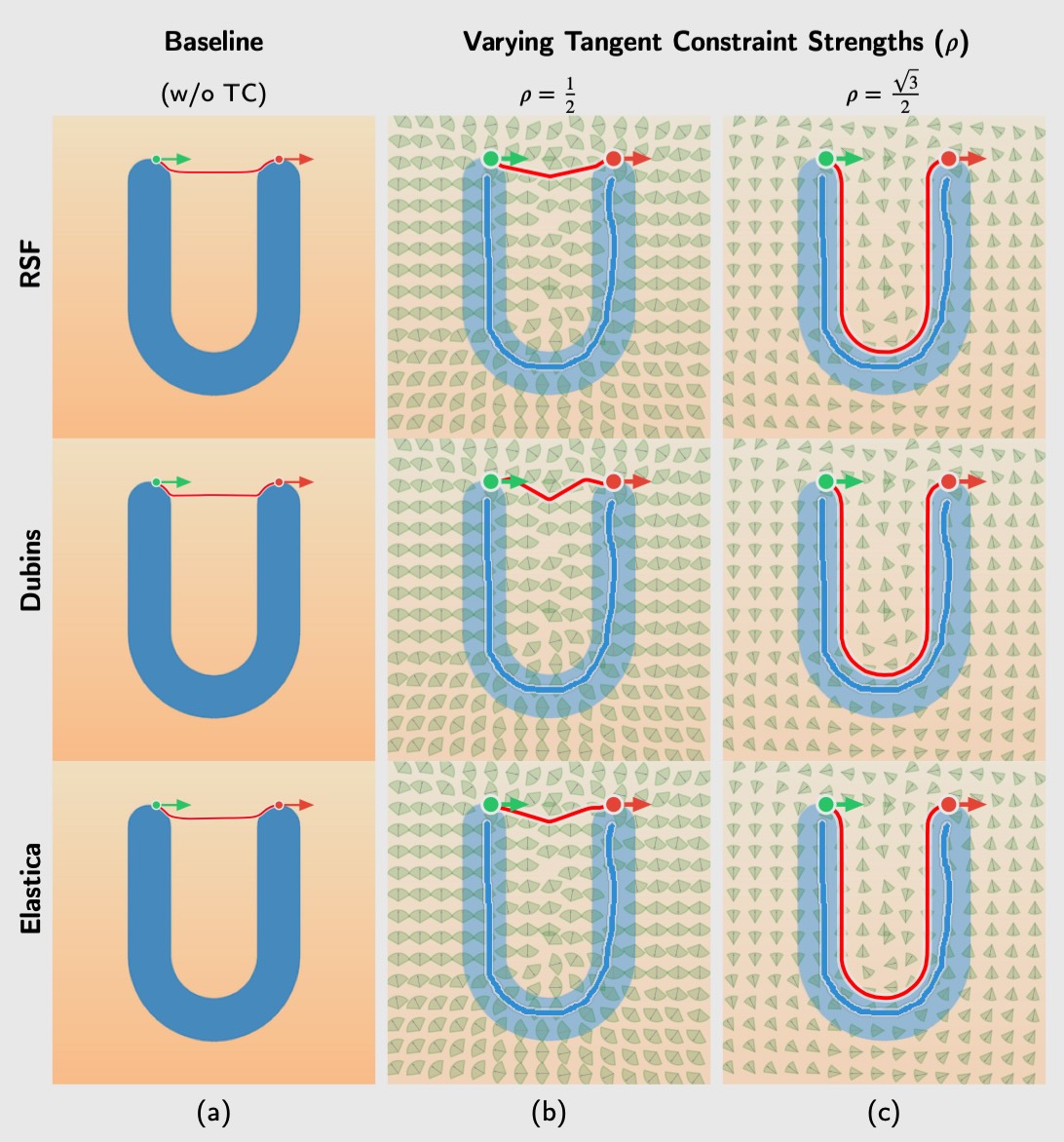}
\caption{Qualitative comparison of different geodesic models on a hairpin shape with varying levels of tangent constraints (TC). The top, middle, and bottom rows show results for the RSF, Dubins, and Elastica models, respectively. Column (a) displays results without tangent constraints. Columns (b) and (c) show results with varying tangent constraint strengths, obtained by setting $\rho$ in \eqref{eq_TangentConstraint} to $\frac{1}{2}$ and $\frac{\sqrt{3}}{2}$, corresponding to acute angles of $\pi/3$ and $\pi/6$, respectively.}
  \label{fig:varying_tangent_constraints}
\end{figure}

\begin{figure*}[htbp]
  \centering
  \includegraphics[width=0.7\linewidth]{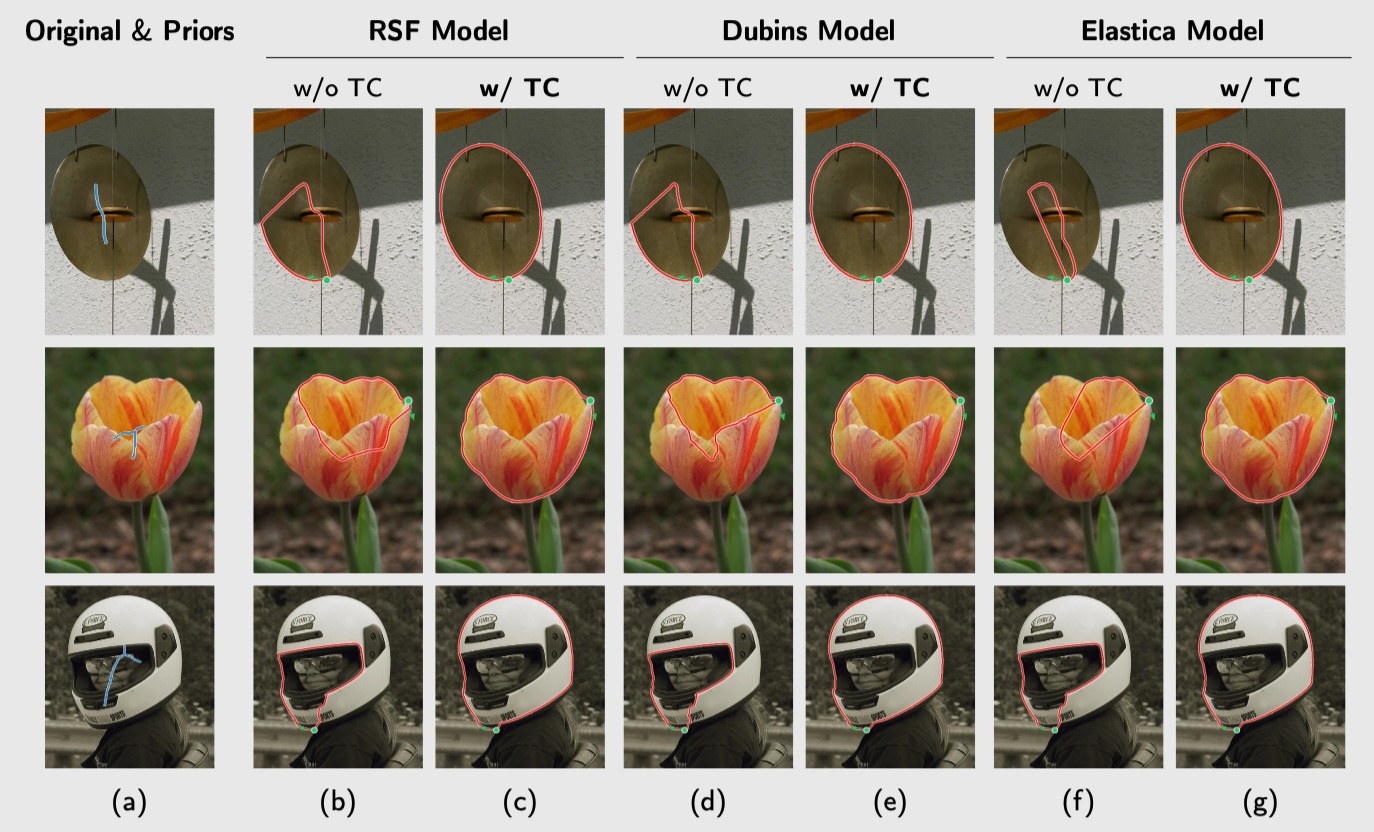}  
  \caption{Qualitative comparison of results obtained with and without tangent constraints. (a) Original images and priors. (b)-(g) Segmentation results for RSF, Dubins, and Elastica models without (w/o) and with (w/) tangent constraints, respectively.}
  \label{fig:effect_of_tangent_priors}
\end{figure*}

In Fig.~\ref{fig:varying_tangent_constraints}, we evaluate the influence of tangent constraint strictness on the minimal paths generated by the RSF, Dubins, and Elastica models. The strictness is governed by the scalar parameter $\rho$ in the acute alignment condition (Eq.~\eqref{eq_TangentConstraint}), where increasing $\rho$ from $0$ to $1$ narrows the admissible angular sector centered on the tangent prior $\kq$. 
Fig.~\ref{fig:varying_tangent_constraints} compares the baseline models (without tangent constraints, column (a)) against their tangent-constrained counterparts with maximum angular deviations of $\pi/3$ and $\pi/6$, as illustrated in columns (b) and (c). In the absence of constraints or when constraints are loose (column (b)), all models suffer from severe shortcutting, failing to track the hairpin contour and instead cutting directly across the gap between the start (green) and end (red) points. 
As the constraints become stricter, the paths are forced to align better with the boundary; notably, with a $\pi/6$ constraint, all models successfully recover the true contour.

We further evaluate the impact of tangent-constrained priors on image segmentation using the proposed framework. Fig.~\ref{fig:effect_of_tangent_priors} provides a qualitative comparison between standard curvature-penalized models (RSF, Dubins, Elastica) and their tangent-constrained counterparts (TC-RSF, TC-Dubins, TC-Elastica) on images from the MSRA-B dataset~\citep{WangDRFI2017}. 
In the first two rows, the objects exhibit weak internal image gradients, resulting in low values of the data-driven cost $\psi$ within the object interior. Consequently, the unconstrained geodesic models (columns (b), (d), (f)) tend to take shortcuts through the object's body to minimize the total energy.
In contrast, by incorporating the tangent constraint into the metric $\kF_{\rho,\kq}$ (Eq.~\eqref{eq_ConstrainedMetric1}), the proposed models restrict the admissible path tangents to align with $\kq$. This effectively prevents the geodesics from traversing the object interior, forcing them to follow the boundary even where the image edge evidence is weak (columns (c), (e), (g)).
The third row illustrates a scenario where the object contains salient internal structures that act as local traps. Without tangent constraints, the geodesics are distracted by these internal boundaries, failing to encompass the entire object. However, the tangent priors successfully guide the optimization, ensuring the recovered paths respect the global topology of the object and avoid getting trapped in local minima.

\begin{figure*}[t]
  \centering
  \includegraphics[width=0.8\linewidth]{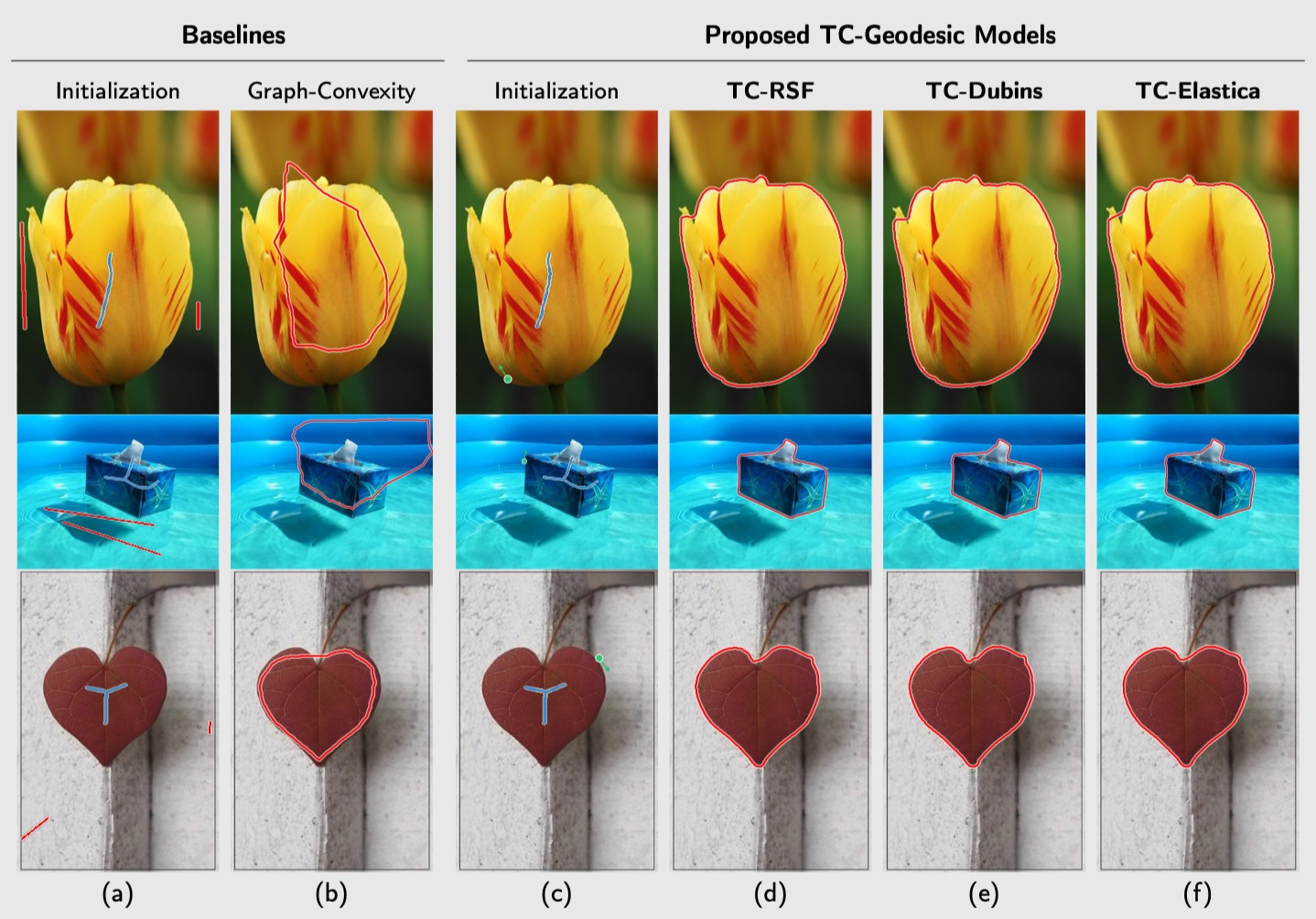}
  \caption{Qualitative comparison between Graph-Convexity and the proposed tangent-constrained geodesic models. Column (a) shows the initialization for Graph-Convexity, and column (b) presents its corresponding segmentation result. Column (c) illustrates the initialization for the geodesic-based models, while columns (d), (e), and (f) show the segmentation results obtained using the TC-RSF, TC-Dubins, and TC-Elastica models, respectively.}
  \label{fig:interactive_comparison}
\end{figure*}

\begin{table*}[htbp]
\centering
\caption{Quantitative evaluation of interactive segmentation on the MSRA-B dataset. Mean and standard deviation of the Dice coefficient are reported.}
\label{tab:interactive_msrab}
\setlength{\tabcolsep}{5pt}
\begin{tabular*}{\textwidth}{@{\extracolsep{\fill}}lccccccc}
\toprule
\textbf{Metric}
& \textbf{Graph-Convexity} & \textbf{RSF} & \textbf{Dubins} & \textbf{Elastica}
& \textbf{TC-RSF} & \textbf{TC-Dubins} & \textbf{TC-Elastica} \\
\cmidrule(lr){2-2}\cmidrule(lr){3-5} \cmidrule(lr){6-8}
\textbf{Mean}
&  0.7359 & 0.7790 & 0.6893 & 0.7157
& \textbf{0.9817} & \textbf{0.9829} & \textbf{0.9799} \\
\textbf{Std}
&0.1650 & 0.3064 & 0.3539 & 0.3490
& \textbf{0.0121} & \textbf{0.0106} & \textbf{0.0155} \\
\bottomrule
\end{tabular*}
\end{table*}

\subsection{Evaluation on Interactive Segmentation}

We evaluate the proposed tangent-constrained geodesic models on the interactive image segmentation task. In this setting, the priors required for segmentation are provided through user interaction. Specifically, for our method, the user provides the skeleton (serving as the ISR to generate the tangent priors $\kq$ via Eq.~\eqref{eq_tangentPriors}) along with the source point $\fp$ and its tangent direction $\theta_p$ for the circular geodesic model, as described in Sec.~\ref{subsec_Circular}.

For comparison, we select the graph-based Convexity Shape Prior (Graph-Convexity) method~\citep{gorelick2016convexity}, a representative shape-aware interactive segmentation approach that enforces global convexity constraints through graph cuts optimization. This method requires user-provided scribbles indicating both foreground and background regions. To ensure a fair comparison, we set the foreground scribble identical to the skeleton used by our method, thereby sharing the same interior shape representation.

Fig.~\ref{fig:interactive_comparison} presents a qualitative comparison between Graph-Convexity and the proposed tangent-constrained geodesic models (TC-RSF, TC-Dubins, TC-Elastica). The results reveal that Graph-Convexity, constrained by limited user input, frequently fails to capture the true object contour---either under-segmenting parts of the object, including extraneous background regions, or producing imprecise boundary delineation. In contrast, our geodesic models leverage the tangent-constrained metric $\kF_{\rho,\kq}$ (Eq.~\eqref{eq_ConstrainedMetric1}) to restrict admissible path tangents, enabling accurate segmentation even with minimal user guidance.

We further conduct quantitative evaluation using the Dice similarity coefficient (DSC):
\begin{equation}
  \label{eq:dice}
\mathrm{Dice}(S, GT) = \frac{2 \, |S \cap GT|} {|S| + |GT|},
\end{equation}
where $S$ and $GT$ denote the segmentation result and the ground truth, respectively.
Experiments are performed on 20 images from the MSRA-B dataset~\citep{WangDRFI2017}, with results summarized in Tab.~\ref{tab:interactive_msrab}. The proposed tangent-constrained models achieve substantially higher mean Dice scores (TC-RSF: 0.9817, TC-Dubins: 0.9829, TC-Elastica: 0.9799) compared to both Graph-Convexity (0.7359) and the unconstrained geodesic models (RSF: 0.7790, Dubins: 0.6893, Elastica: 0.7157). Notably, the standard deviations of our models are an order of magnitude smaller, indicating consistently reliable performance across diverse object shapes. The unconstrained geodesic models exhibit high variance due to their susceptibility to shortcutting through low-gradient object interiors, as discussed in the previous subsection. These results demonstrate that incorporating the tangent-constrained priors effectively prevents such degenerate solutions, yielding robust segmentation with minimal user input.

\subsection{Evaluation on Automated Segmentation}

\begin{figure*}[htbp]
  \centering
  \includegraphics[width=0.8\linewidth]{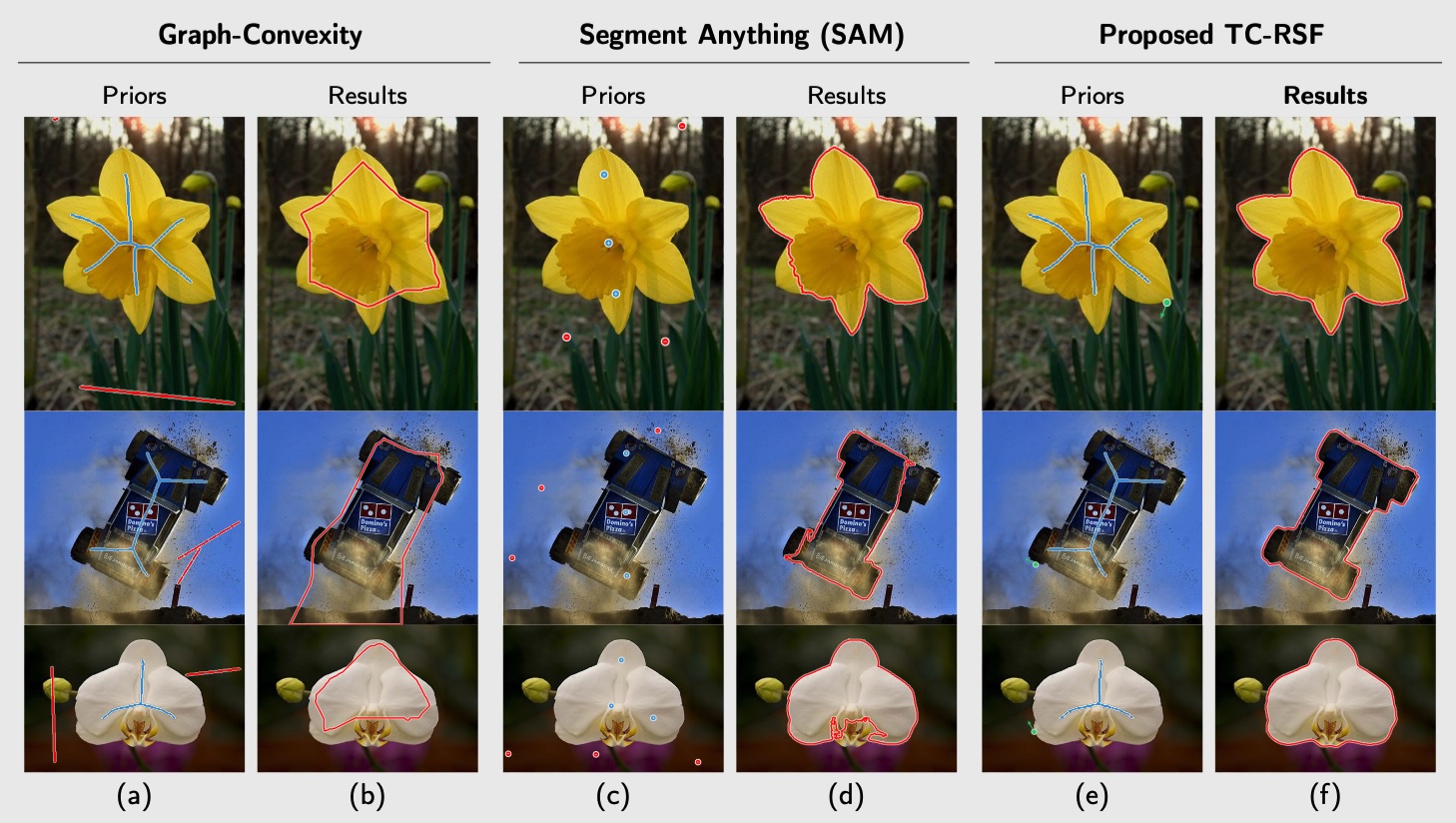}
  \caption{Qualitative comparison on natural images from the MSRA-B dataset in the automated segmentation setting. All methods utilize priors derived from neural network pre-segmentation. Columns (a) and (b) show the priors and segmentation results for Graph-Convexity; (c) and (d) display those for SAM; and (e) and (f) present the priors and results for the proposed TC-RSF model.}
  \label{fig:natural_images_with_priors}
\end{figure*}


\begin{figure*}[t]
  \centering
  \includegraphics[width=0.8\linewidth]{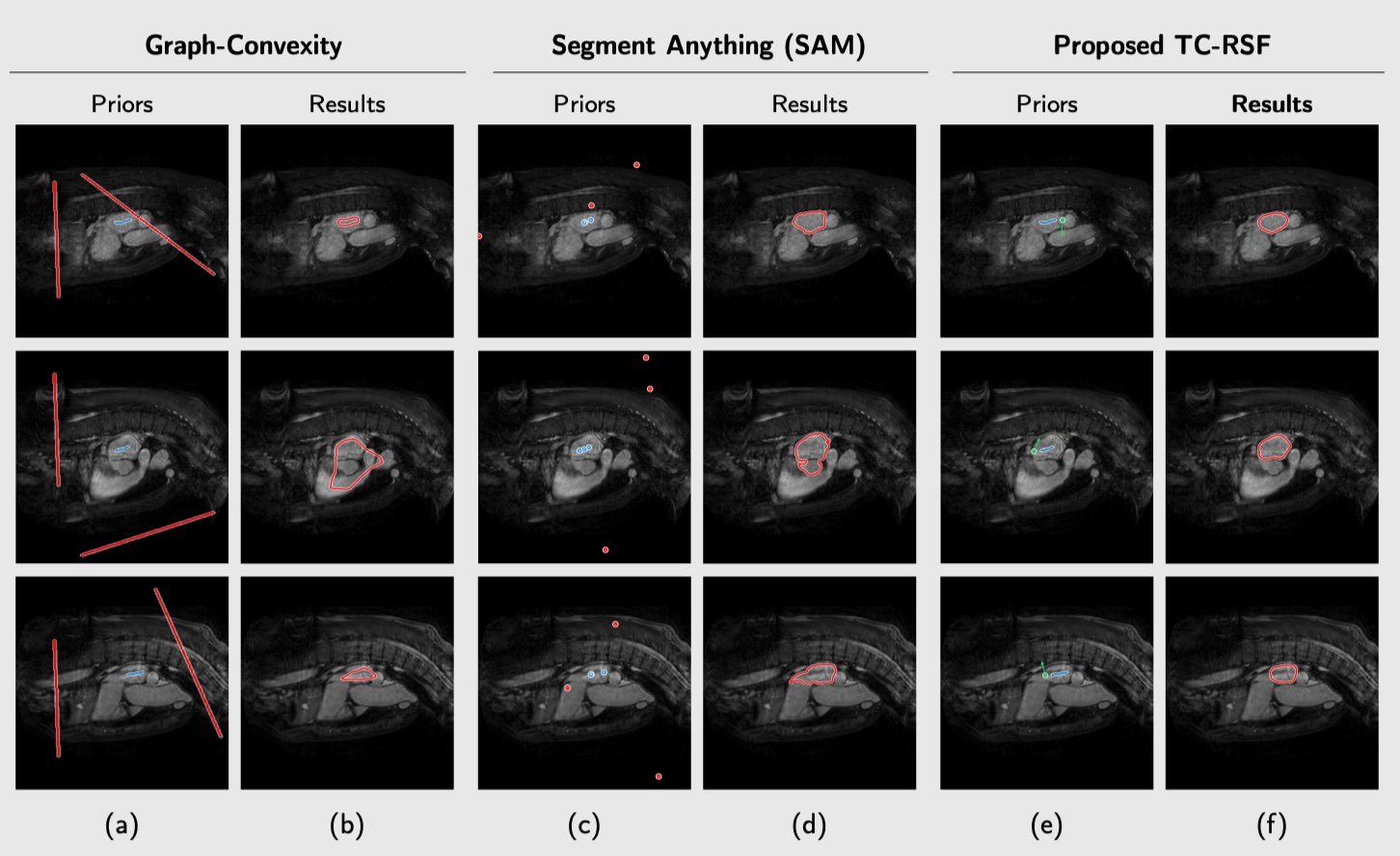}
  \caption{Qualitative comparison on medical images from the MSD Left Atrium dataset in the automated segmentation setting. All methods utilize priors derived from neural network pre-segmentation. Columns (a) and (b) show the priors and segmentation results for Graph-Convexity; (c) and (d) display those for SAM; and (e) and (f) present the priors and results for the proposed TC-RSF model.}
  \label{fig:medical_images_with_priors}
\end{figure*}

\begin{table*}[htbp]
\centering
\caption{Quantitative evaluation of automated segmentation on the MSRA-B (natural images) and MSD Left Atrium (medical images) datasets. Mean and standard deviation of the Dice coefficient are reported.}
\label{tab:mean_std_results}
\begin{tabular*}{0.8\textwidth}{@{\extracolsep{\fill}}lcccc}
\toprule
\textbf{Method}
& \multicolumn{2}{c}{\textbf{Natural Images (MSRA-B)}}
& \multicolumn{2}{c}{\textbf{Medical Images (MSD LA)}} \\
\cmidrule(lr){2-3} \cmidrule(lr){4-5}
& \textbf{Mean} & \textbf{Std}
& \textbf{Mean} & \textbf{Std} \\
\midrule
DeepLabv3+            & 0.9407 & 0.0416 & 0.8877 & 0.0678 \\
SAM                   & 0.8103 & 0.2271 & 0.7926 & 0.1998 \\
Graph-Convexity       & 0.8008 & 0.1482 & 0.4258 & 0.1156 \\
RSF                   & 0.9144 & 0.1755 & 0.8690 & 0.1259 \\
Dubins                & 0.8947 & 0.2064 & 0.8266 & 0.1538 \\
Elastica              & 0.8807 & 0.2156 & 0.8832 & 0.0999 \\
\midrule
\textbf{TC-RSF}
                       & \textbf{0.9732} & \textbf{0.0247}
                       & \textbf{0.9307} & \textbf{0.0373} \\
\textbf{TC-Dubins}
                       & \textbf{0.9739} & \textbf{0.0192}
                       & \textbf{0.9173} & \textbf{0.0601} \\
\textbf{TC-Elastica}
                       & \textbf{0.9681} & \textbf{0.0469}
                       & \textbf{0.9301} & \textbf{0.0394} \\
\bottomrule
\end{tabular*}
\end{table*}

In this subsection, we evaluate the proposed framework in the automated segmentation setting, where all priors are derived from neural network pre-segmentation without user interaction. As described in Sec.~\ref{sec:pre-segmentation}, a DeepLabv3+ network provides the initial segmentation from which we extract the skeleton $\cS$ (serving as the ISR) to construct the tangent priors $\kq$ and initialize the CGM.

We compare against two representative methods: the graph-based Graph-Convexity~\citep{gorelick2016convexity} and the Segment Anything Model (SAM)~\citep{kirillov2023segany}. To ensure fair comparison, we adapt these originally interactive methods to utilize the same pre-segmentation. Specifically, for Graph-Convexity, the foreground scribbles are derived from the skeleton $\cS$. For SAM, positive prompt points are sampled from this skeleton. Since both baseline methods require background information (unlike our approach, which only requires the ISR), we generate background scribbles (for Graph-Convexity) and negative prompt points (for SAM) by randomly sampling from the region external to the pre-segmentation mask. All geodesic-based models share identical initialization parameters, including the source point $\fp$ and the cut configuration for the CGM.

Fig.~\ref{fig:natural_images_with_priors} presents qualitative results on natural images from the MSRA-B dataset. The convexity constraint in Graph-Convexity proves overly restrictive, failing to capture non-convex boundary details when object shapes are complex (columns (a)--(b)). SAM, while capable of handling complex topologies, often struggles with precise boundary delineation, leading to under-segmentation or loss of fine structural details (columns (c)--(d)). In contrast, our TC-RSF model (columns (e)--(f)) leverages the tangent-constrained metric $\kF_{\rho,\kq}$ (Eq.~\eqref{eq_ConstrainedMetric1}) combined with curvature regularization, enabling precise contour tracking even for objects with intricate shapes.

We further evaluate on medical images using the Left Atrium dataset from the Medical Segmentation Decathlon (MSD)~\citep{antonelli2022medical}. As shown in Fig.~\ref{fig:medical_images_with_priors}, the low contrast and noise inherent in medical imaging pose significant challenges. Graph-Convexity struggles to localize boundaries accurately despite the target's relatively convex shape, likely due to weak edge evidence in the data-driven cost $\psi$. SAM tends toward over-segmentation, erroneously including adjacent tissues due to ambiguous intensity boundaries. Our TC-RSF model demonstrates superior robustness by leveraging the tangent priors $\kq$ to guide the geodesic through low-contrast regions where image gradients alone are insufficient.

Table~\ref{tab:mean_std_results} presents the quantitative results on both datasets, computed using 100 selected representative images. The proposed tangent-constrained models consistently achieve the highest Dice scores: TC-Dubins leads on natural images (0.9739) while TC-RSF leads on medical images (0.9307). Compared to their unconstrained counterparts, the tangent-constrained models yield improvements of 5--9\% (e.g., TC-RSF vs.\ RSF: 5.9\% on MSRA-B, 6.2\% on MSD LA), confirming that the tangent priors $\kq$ effectively prevent shortcut paths. The standard deviations are also reduced by an order of magnitude (e.g., TC-Dubins: 0.0192 vs.\ Dubins: 0.2064), reflecting the stability provided by the hard constraint $\kC_{\rho,\kq}$ (Eq.~\eqref{eq_AcuteConstraint}). Notably, all three tangent-constrained models outperform the DeepLabv3+ network that provides their initialization, demonstrating that the geometric constraints in the HJB PDE formulation (Eq.~\eqref{eq_HJB_ours}) effectively refine neural network predictions. In contrast, Graph-Convexity and SAM perform poorly, particularly on medical images (0.4258 and 0.7926, respectively), due to overly restrictive convexity assumptions and lack of domain-specific training.

\section{Conclusion}
\label{sec_Conclusion}
In this work, we proposed a unified framework that incorporates tangent constraints as spatially varying shape constraints into curvature-regularized geodesic models within the HJB PDE formulation. By enforcing tangent admissibility in the orientation-lifted domain, our model restricts geodesic trajectories to prescribed angular sectors while remaining fully compatible with curvature penalization. The resulting tangent-constrained metric can be efficiently optimized using an adapted Hamiltonian Fast Marching algorithm, enabling the extraction of circular geodesics and closed contours for image segmentation. Experimental results on natural and medical images confirm that the proposed approach improves robustness under weak or noisy boundaries, mitigates topological shortcuts, and achieves competitive or superior segmentation performance compared to deep learning baselines and classical geodesic models without priors.

Future work will be devoted to extending the framework to 3D volumetric segmentation in conjunction with the framework~\citep{chambolle2019total}, and developing end-to-end learning pipelines for joint extraction of tangent priors and the circular geodesic computation.

\appendix
\section{Curvature-penalized Geodesic Models}
\label{appendix_CurvatureModels}
Existing second-order geodesic models~\citep{chen2017global,chen2023computing,mirebeau2018fast,duits2018optimal,van2024geodesic} employ a curvature-dependent regularization term to construct the corresponding bending energy functional, see~\cref{eq:energy}. Typically, those models differ from each other mainly in the curvature cost $\cC$ used. Specifically, the cost $\cC$ for the Reeds-Shepp forward (RSF) model~\citep{duits2018optimal}, the Dubins (D) model~\citep{mirebeau2018fast} and the Euler-Mumford elastica (EM) model~\citep{chen2017global} can be respectively formulated as
\begin{align*}
&\cC_{\rm RSF}(a)=\sqrt{1+(a)^2},\quad 
\cC_{\rm D}(a)=
\begin{cases}
1,&\text{if~}|a|\leq 1\\
\infty,&\text{otherwise}
\end{cases},\\
&\cC_{\rm EM}(a)=1+(a)^2,
\end{align*}
for any scalar value $a\in\bR$.

In addition, the curvature prior elastica (CP) model~\citep{chen2023computing} invokes the same curvature cost as the EM elastica model, except that the CP model utilizes a shifted curvature term $\tilde\kappa$ defined such that 
\begin{equation*}
\tilde\kappa(t)=\kappa(t)-\varsigma(\gamma(t),\eta(t)),\quad \forall t\in[0,1],	
\end{equation*}
where $\kappa$ is the curvature of a regular curve $\gamma$, and $\varsigma:\bM\to\bR$ is a data-driven function which is also referred to as curvature priors. For a point $x$ located in a centerline of tubular structure, $\varsigma(x,\theta)$ is the estimated curvature of a fitting curve (to that centerline) whose path tangent is $\rn(\theta)$. We refer the reader to~\citep{chen2023computing} for the computation of curvature priors $\varsigma$.

\section{Orientation Scores from Image Gradients}
\label{appendix_OSGrad}
Image edge and anisotropy features are extracted from image gradients. To enhance robustness to noise, the gradients are computed from a smoothed version of the image. First, the image $I$ is smoothed by convolving it with a Gaussian kernel $G_\sigma$ with standard deviation $\sigma$, resulting in $G_\sigma * I$. The core of this method is the Jacobian matrix, $\cJ$, which contains the partial derivatives of the smoothed image. For a color image $I = (I_1, I_2, I_3)$, the Jacobian at a point $x\in\Omega$ is a matrix of size $2 \times 3$:
\begin{equation}
\cJ(x) = 
\begin{pmatrix} 
\partial_a G_\sigma * I_1 & \partial_a G_\sigma * I_2 &\partial_a G_\sigma * I_3 \\ 
\partial_b G_\sigma * I_1 & \partial_b G_\sigma * I_2 & \partial_b G_\sigma * I_3 
 \end{pmatrix},
\end{equation}
where $\partial_a G_\sigma$ (resp. $\partial_b G_\sigma$) is the differential of the kernel $G_\sigma$ along the axis $a$ (resp. axis $b$). 

For a grayscale image, the Jacobian simplifies to a $2 \times 1$ vector of the smoothed image's partial derivatives:
\begin{equation}
  \cJ(x) = (\partial_a G_\sigma * I, \partial_b G_\sigma * I)^\top(x).
\end{equation}
The anisotropic features, describing the orientation and shape of edges, are encapsulated in a tensor field $\cW$. At each point $\mathbf{x}$, $\cW(\mathbf{x})$ is a $2 \times 2$ symmetric positive-definite matrix
\begin{equation*}
 \cW(x) = \cJ(x)\cJ(x)^\top + \Id
\end{equation*}
where $\Id$ is the identity of size $2\times2$, ensuring that the matrix $\cW(x)$ is positive-definite for any point $x\in\Omega$.

The image-driven cost function $\psi_{\rm grad}$ is formulated to guide the minimal path towards salient image edges. It incorporates the anisotropic feature from $\cW$:
\begin{equation}
\label{eq_gradient_cost}
\psi_{\rm grad}(\fx) =\frac{\left\langle \rn(\theta)^{\perp}, \cW(x) \rn(\theta)^{\perp} \right\rangle^{1/2}}{\max_{(y,\vartheta)\in\bM}\left\langle \rn(\vartheta)^{\perp}, \cW(x) \rn(\vartheta)^{\perp} \right\rangle^{1/2}}
\end{equation}
for any point $\fx=(x,\theta) \in \bM$, where $\rn(\theta) = (\cos\theta, \sin\theta)$ is a unit vector associated with angle $\theta$, $\rn(\theta)^{\perp} = (-\sin\theta, \cos\theta)$ is its perpendicular vector, and $\langle \cdot, \cdot \rangle$ denotes the standard Euclidean scalar product. The term $\left\langle \rn_{\theta}^{\perp}, \cW(x) \rn_{\theta}^{\perp} \right\rangle$ serves as an orientation score. This score is maximized when the vector $\rn_{\theta}^{\perp}$ is aligned with the image gradient direction which is the principal direction of $\cW(x)$. This, in turn, encourages the path's tangent vector, $\rn_{\theta}$, to align with the orientation of the edge, where the cost $\psi_{\rm grad}(\mathbf{x})$ is minimal. This design is advantageous for segmentation, as it generates a valley of low potential along image edges, thereby allowing to attract the minimal paths to pass through the desired edge points.

\section{Orientation Scores from Region-based Appearance Models}
\label{appendix_RegionalScores}

Beyond image gradient features, region-based image appearance features provide complementary cues for guiding geodesic paths for delineating the target boundary enclosing regions with complicated image content. We adopt the region competition model with Gaussian mixture model (GMM) as an example for quantifying the likelihood of image intensities belonging to foreground or background regions. Note that other region-based appearance models such as image color histograms-based models can be considered, and we refer the reader to~\citep{chen2024region,chen2023geodesic} for more detail. 

Consider a simple closed curve $\gamma:[0,1]\to\Omega$ that divides the domain $\Omega$ into two disjoint regions: $\cR_1$ (interior) and $\cR_2$ (exterior) with $\Omega=\cR_1\cup \cR_2$. For each region, we define a data fidelity function $\rho_i:\Omega\to\bR$ (for $i=1,2$) based on a Gaussian mixture model (GMM), where $\rho_i(x)=-\log\,P_i(I(x)~|~\Theta_i)$ represents the negative log-likelihood of the image intensity $I(x)$ under the probability density function $P_i$ with the parameters $\Theta_i$.

To incorporate these regional statistics into the geodesic framework, we utilize the method introduced in~\citep{chen2024region} to encode the difference $\rho_1-\rho_2$ into a vector field $\omega:\bR^2\to\bR^2$ of sufficiently small magnitude. 
This is implemented by addressing the following PDE-constrained minimization problem
\begin{equation}
\min~\int_{\bR^2}\|\omega(x)\|^2\,dx,~s.t.~\curl\omega=(\rho_1-\rho_2)\chi_U,
\end{equation}
 where $\chi_U\subset\Omega$ denotes the indicator function of a bounded tubular neighbourhood $U \subset \Omega$. In other words, among all feasible solutions to the PDE $\curl\omega=(\rho_1-\rho_2)\chi_U$, we choose the one minimizing the energy $\int_{\bR^2}\|\omega(x)\|^2dx$, which admits a closed-form solution via convolution.

\bibliographystyle{plainnat}

\bibliography{cas-refs}



\end{document}